\documentclass{article}
\usepackage[top=1in, bottom=1in, left=1in, right=1in]{geometry}
\usepackage{tgpagella}
\usepackage[utf8]{inputenc}
\usepackage[T1]{fontenc}

\usepackage{url}
\usepackage[hidelinks]{hyperref}
\usepackage{amsmath,amsthm,amssymb,amsbsy,bm}
\usepackage{paralist}
\usepackage{xcolor}
\usepackage{color}
\usepackage{graphicx}
\graphicspath{{./figs/}}
\usepackage{algorithm}
\usepackage{algorithmic}
\usepackage{comment}
\usepackage{multirow}
\usepackage{enumerate}
\usepackage{enumitem}
\usepackage{graphicx}
\usepackage{fancyhdr}
\usepackage{cite}
\usepackage{cleveref}
\usepackage{subcaption}
\newtheorem{lemma}{Lemma}%[section] %%    with section number.
\theoremstyle{remark}

\theoremstyle{problem}

\newcommand{\R}{\mathbb{R}}

\newcommand{\e}{\begin{equation}}
\newcommand{\ee}{\end{equation}}
\newcommand{\en}{\begin{equation*}}
\newcommand{\een}{\end{equation*}}
\newcommand{\eqn}{\begin{eqnarray}}
\newcommand{\eeqn}{\end{eqnarray}}
\newcommand{\bmat}{\begin{bmatrix}}
\newcommand{\emat}{\end{bmatrix}}
\DeclareMathAlphabet\mathbfcal{OMS}{cmsy}{b}{n}
\newcommand{\E}{\operatorname{\mathbb{E}}}

\newcommand{\vct}[1]{\boldsymbol{#1}}
\newcommand{\mtx}[1]{\boldsymbol{#1}}
\newcommand{\set}[1]{\mathbb{#1}}
\DeclareMathOperator*{\argmin}{\text{arg~min}}

\newcommand{\wh}{\widehat}

\newcommand{\calH}{\mathcal{H}}

\newcommand{\calT}{\mathcal{T}}

\newcommand{\vb}{\vct{b}}

\newcommand{\vd}{\vct{d}}

\newcommand{\vh}{\vct{h}}

\newcommand{\vu}{\vct{u}}
\newcommand{\vv}{\vct{v}}

\newcommand{\vx}{\vct{x}}

\newcommand{\vz}{\vct{z}}
\newcommand{\vzero}{\vct{0}}

\newcommand{\mD}{\mtx{D}}

\newcommand{\mW}{\mtx{W}}
\newcommand{\mX}{\mtx{X}}

\newcommand{\setA}{\set{A}}

\graphicspath{{./figs/}}

\newlength{\imgwidth}
\newboolean{twoColVersion}
\setboolean{twoColVersion}{false}
\newcommand{\twoCol}[2]{\ifthenelse{\boolean{twoColVersion}} {#1} {#2} }

\usepackage{url}
\usepackage{booktabs}
\usepackage{hyperref}
\hypersetup{
    colorlinks=true,
    linkcolor=blue,
    citecolor=blue,
    urlcolor=blue
}
\usepackage{fontawesome5}

\title{AbsTopK: Rethinking Sparse Autoencoders For Bidirectional Features}
\author{
Xudong Zhu, Mohammad Mahdi Khalili, Zhihui Zhu
  \\  \vspace{1ex}
Department of Computer Science \& Engineering, The Ohio State University \\ \vspace{1ex}
  \;\texttt{\{zhu.3944,\;\;khaliligarekani.1,\;\;zhu.3440\}@osu.edu} \\
  \faGithub\hspace{0.3em}\href{https://github.com/GoXzascc/LLMInterpretabilityHub.git}{AbsTopK}
}

\begin{document}

\maketitle

\begin{abstract}
Sparse autoencoders (SAEs) have emerged as powerful techniques for interpretability of large language models (LLMs), aiming to decompose hidden states into meaningful semantic features. While several SAE variants have been proposed, there remains no principled framework to derive SAEs from the original dictionary learning formulation. In this work, we introduce such a framework by unrolling the proximal gradient method for sparse coding. We show that a single-step update naturally recovers common SAE variants, including ReLU, JumpReLU, and TopK. Through this lens, we reveal a fundamental limitation of existing SAEs: their sparsity-inducing regularizers enforce non-negativity, preventing a single feature from representing bidirectional concepts (e.g., male vs. female). This structural constraint fragments semantic axes into separate, redundant features, limiting representational completeness. To address this issue, we propose AbsTopK SAE, a new variant derived from the $\ell_0$ sparsity constraint that applies hard thresholding over the largest-magnitude activations. By preserving both positive and negative activations, AbsTopK uncovers richer, bidirectional conceptual representations. Comprehensive experiments across four LLMs and seven probing and steering tasks show that AbsTopK improves reconstruction fidelity, enhances interpretability, and enables single features to encode contrasting concepts. Remarkably, AbsTopK matches or even surpasses the Difference-in-Mean method—a supervised approach that requires labeled data for each concept and has been shown in prior work to outperform SAEs.

\end{abstract}

\section{Introduction}

The pursuit of interpretability has become a central objective in modern machine learning, as it is essential for the assurance, debugging, and fine-grained control of large language models (LLMs) \cite{marks2025sparse, park2023linear, luo2024pace, arora-etal-2018-linear}. Within this domain, sparse dictionary learning methods \cite{Poggio2006LearningAD, fel2023craft}, and specifically sparse autoencoders (SAEs), have re-emerged as a prominent methodology for systematically enumerating the latent concepts a model may employ in its predictions \cite{hindupur2025projecting, bussmann2024batchtopk, rajamanoharan2025jumping, topk}.

An SAE decomposes a model's hidden representations into an overcomplete basis of latent features \cite{elhage2022toy, thasarathan2025universal}, which ideally correspond to abstract, data-driven concepts whose linear superposition reconstructs the original activation vector \cite{higgins2017betavae, Fel2025SparksOE}. Empirical evidence indicates that SAE latents capture semantically coherent features across diverse domains. In LLMs, these features exhibit selectivity for specific entities (e.g., \emph{Golden Gate Bridge}), linguistic behaviors (e.g., sycophantic phrasing), and symbolic systems (e.g., Hebrew script) \cite{templeton2024scaling, csordas-etal-2024-recurrent, durmus2024steering}. Similarly, in vision models, they respond to distinct objects (e.g., barbers, dog shadows) and complex scene properties (e.g., foreground-background separation, facial detection in crowds) \cite{fel2024sparks, thasarathan2025universal}. In protein models, they have been shown to correlate with functional elements such as binding sites and structural motifs \cite{garcia2025interpreting, Adams2025bio}. The discovery of such interpretable, semantically grounded features suggests a natural avenue for steering models: by amplifying, suppressing, or combining specific latents, one can intervene to modulate downstream behavior, which is a principal motivation for research into SAEs \cite{topk, bricken2023monosemanticity, kantamneni2025are}. This control is predicated on the assumption that the concepts identified by SAEs faithfully correspond to the features underlying a model's predictions \cite{arditi2024refusal, uppaal2024model, engels2025not}.

However, recent studies suggest that simpler supervised techniques such as Difference-in-Means (DiM) can outperform SAEs on practical steering benchmarks and tasks \cite{arditi2024refusal, wu2025axbench}. Unlike SAEs, which are unsupervised and can simultaneously identify multiple latent features, DiM requires labeled data and is typically limited to extracting a single vector for a pre-specified concept. Nevertheless, these findings raise questions regarding the degree to which SAEs recover a model’s internal features. The fact that comparatively simple baselines can rival or even surpass SAEs on downstream control tasks suggests that the features identified by SAEs may only partially align with the model’s underlying neural representations, thereby casting doubt on their fidelity as faithful explanatory tools.

\begin{figure}[t]
    \centering
    \includegraphics[width=\linewidth]{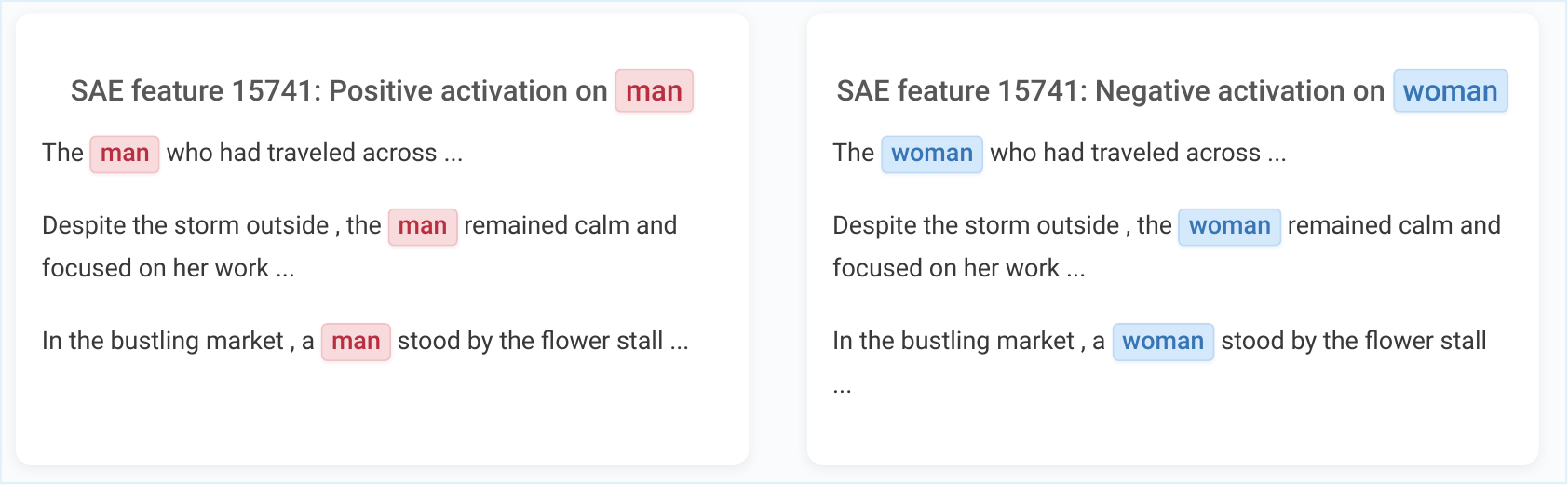}
    \caption{\textbf{AbsTopK enables single latent features to encode opposing concepts by leveraging both positive and negative activations.}
    To test this, we generated controlled sentence pairs with only one differing token (\emph{man} vs.\ \emph{woman}).
    The shown feature activates positively for \emph{man} and negatively for \emph{woman}, demonstrating bidirectional encoding.
    Unlike conventional SAEs, which are restricted by a non-negativity constraint, AbsTopK more compactly captures opposing semantics within a single dimension, yielding richer and more coherent representations.}
    \label{fig:feature_example}
\end{figure}

We posit that one source of this misalignment lies in a structural limitation of SAEs recently proposed for studying LLMs—including the vanilla version with ReLU \cite{cunningham2023sparse}, the JumpReLU variant \cite{rajamanoharan2025jumping}, and the TopK variant \cite{topk}: their systematic neglect of negative activations, despite evidence that many meaningful directions in representation space are inherently bidirectional \cite{Mao2022BidirectionalFG}. The \emph{linear representation hypothesis} \cite{Mikolov2013EfficientEO} suggests that a model's internal states can be approximated as linear combinations of semantic vectors, where conceptual transformations correspond to both positive and negative displacements along these vector axes \cite{arora-etal-2018-linear, uppaal2024model, luo2024pace}. The DiM approach builds on this assumption, requiring labeled datasets that capture both sides of a concept, with positive and negative examples defining a bidirectional semantic axis. Classic word analogies, such as the vector operation $\vv_{\text{king}} - \vv_{\text{man}} + \vv_{\text{woman}} \approx \vv_{\text{queen}}$ \cite{pennington-etal-2014-glove}, illustrate how semantic differences are encoded as generalizable vector offsets. Nevertheless, by enforcing non-negativity or retaining only the TopK  activations \cite{bussmann2024batchtopk, topk}, conventional SAEs either fragment such contrastive concepts into separate, unidirectional bases (e.g., ``male" and ``female") or discard one direction of the semantic axis entirely. This not only undermines the representational capacity of SAEs but also limits their usefulness for controlled interventions, where traversing both directions of a semantic axis is often essential. This raises the following questions: {\it is the use of nonnegative activations truly essential for the success of SAEs, or does it instead constrain their ability to capture richer representations? More concretely, can SAEs be improved by allowing negative activations, thereby enabling the discovery of bidirectional concepts?}

\paragraph{Contributions.} In this work, we address these questions by $(i)$ introducing a unified framework for designing SAEs, $(ii)$ proposing a new variant, AbsTopK SAE, and $(iii)$ conducting comprehensive experiments across four LLMs and seven probing and steering tasks to demonstrate that allowing negative activations further enhances SAEs, yielding improved reconstruction fidelity and greater interpretability.

Our contributions are summarized as follows:
\begin{itemize}
[leftmargin=*]
\item \textbf{A Unified Framework for Designing SAEs.} We introduce a principled framework for designing SAEs by unrolling the proximal gradient method for sparse coding with sparsity-inducing regularizers. A single-step update naturally induces common SAE variants, including ReLU, JumpReLU, and TopK \cite{templeton2024scaling, topk, rajamanoharan2025jumping}. This framework provides a rigorous tool for analyzing their implicit regularizers and identifying shared limitations.

\item \textbf{Absolute TopK (AbsTopK) for Learning Bidirectional Features}
Building on this framework, we propose a new SAE variant, termed absolute TopK (AbsTopK) derived from the vanilla sparsity constraint ($\ell_0$ norm) without a non-negative constraint, which results in a hard-thresholding operator that selects the largest-magnitude activations. By preserving both positive and negative activations, AbsTopK SAE allows a single feature to capture opposing concepts (Figure~\ref{fig:feature_example}), thereby uncovering richer bidirectional representations.

\item \textbf{Comprehensive Empirical Validation.} We conduct a comprehensive empirical evaluation across four LLMs, comparing the proposed AbsTopK SAE with TopK and JumpReLU SAEs on a suite of seven probing and steering tasks, along with three unsupervised metrics. The results demonstrate that AbsTopK outperforms TopK and JumpReLU SAEs, producing representations with higher fidelity and interpretability. Additionally, a case study illustrates that AbsTopK can encode a bidirectional semantic axis within a single latent feature, effectively capturing contrasting concepts. Notably, AbsTopK achieves performance comparable to—or even exceeding—the Difference-in-Mean method, which relies on labeled data and has been shown in prior work to outperform SAEs.
\end{itemize}

\section{From Proximal Interpretations of SAEs to AbsTopK}

\subsection{Preliminaries}
We denote vectors by lowercase bold letters (e.g., $\vx$) and matrices by uppercase bold letters (e.g., $\mX$).
With an input sequence of $N$ tokens, $\mX = \{\vx_1, \ldots, \vx_N\}$, where each $\vx_j$ denotes the embedding of the $j$-th token, the LLM can be viewed as a function $
    f : \mathbb{R}^{d \times N} \to \mathbb{R}^{V \times N},$ where $V$ is the vocabulary size and $f(\mX)$ gives the output logits for all tokens in the sequence.  For our purposes, we abstract away the internal details of $f$ and instead study the representations in the hidden layers. Consider interventions at layer $\ell$ in the residual stream. Supposing that that the model comprises $L$ layers, then $f$ can be decomposed as
\begin{equation}
    f(\mX) \;=\; \phi_{\ell+1:L}\!\big( \phi_{1:\ell}(\mX) \big),
\end{equation}
where $\phi_{1:\ell}(\mX)$ denotes the representation after the first $\ell$ layers and $\phi_{\ell+1:L}$ represents the remaining computation from layer $\ell$ to $L$.
We denote by $\vx^{(\ell)}_j$ the embedding of the residual stream at the $\ell$-th layer corresponding to the $j$-th token of the input sequence $\mX$.
In the following presentation, when the context is clear, we omit the superscript ($\ell$) and the subscript (token index $j$) for notational simplicity, and denote the hidden embedding of a token in a given layer by \(\vx\).

The linear representation hypothesis \cite{park2023linear} assumes that the hidden representation $\vx$ can be expressed as a linear superposition of latent concepts:
\begin{equation}\label{linear_combination}
    \vx \;=\; \sum_{p=1}^P \alpha_p \vh_p \;+\; \text{residual},
\end{equation}
where $\{\vh_p\}_{p=1}^P$ are referred to as concept directions or feature vectors, such as gender or sentiment, $\{\alpha_p\}$ are the corresponding coefficients, and residual term captures approximation error as well as context-specific variation that is not explained by the selected concepts. Since a particular token—although it encodes information from previous tokens in the context—typically contains only a small subset of concepts or features, its representation is expected to be \emph{sparse}; that is, most of the coefficients $\alpha_p$ are zero, resulting in a {\it sparse linear representation}, often simply referred to as a {\it sparse representation}. Importantly, the coefficients $\alpha_p$ are not required to be non-negative. In fact, for binary concepts, the sign of a coefficient is semantically meaningful, indicating opposite directions; for example, it distinguishes whether a contextually appropriate token should be “king” or “queen” (when the context involves a monarch) \cite{park2023linear}.

To find these concept directions or feature vectors, supervised approaches such as the Difference-in-Mean (DiM) method construct labeled datasets for each target attribute. While effective for isolating specific concepts, these methods are inherently limited to predefined features and do not scale to the large number of latent dimensions present in LLM representations. In contrast, dictionary learning provides an unsupervised and scalable alternative: it can simultaneously recover a more complete dictionary that approximates the underlying concept directions, uncovering a richer and more comprehensive set of latent features than DiM, which is typically restricted to a single concept vector. Consequently, while DiM may achieve stronger control on a specific concept \cite{wu2025axbench}, dictionary-learning methods have gained popularity due to their ability to uncover a richer, more comprehensive set of latent features.

\subsection{Dictionary Learning and the Proximal Perspective on Sparse Autoencoders}

In a nutshell, dictionary learning \cite{olshausen1996emergence} seeks to construct a dictionary $\mD$ consisting of basis vectors $\{\vd_1, \ldots, \vd_{P}\}$, which are also called as {\it atoms}, such that it can (approximately) provide sparse linear combination for all token embeddings $\vx$ from the same layer. Since the total number of concept vectors $P’$ is unknown, $P$ is typically set to a relatively large value to ensure that as many concepts as possible can be learned. This typically requires solving a training problem of form  \cite{mairal2011task,sulam2022recovery,MOD,KSVD}

\begin{equation}
\min_{\mD\in \R^{d\times P},\vb \in \R^d}
    \E_{\vx}\left[\min_{\vz\in\R^s} \underbrace{\frac{1}{2}\left\| \vx - (\bm D\vz + \vb) \right\|_2^2}_{g(\vz)}
    + \lambda R(\vz) \right],
\label{eq:DL-single-level}\end{equation}
 where $R(\vz)$ is a sparsity-inducing regularizer, $\lambda > 0$ controls the trade-off between reconstruction fidelity and sparsity, $\vb$ is an additional bias vector. In classical dictionary learning, the data is often preprocessed to have zero global mean, so the bias term is not used. Alternatively, the bias term can be incorporated into the dictionary as $\mD\vz + \vb = \begin{bmatrix} \mD & \vb \end{bmatrix}\begin{bmatrix}\vz \\ 1\end{bmatrix}$. In this work, however, we explicitly include $\vb$ to align with the structure of commonly used SAEs which will be described later.

The main challenge in solving the problem \eqref{eq:DL-single-level} lies in jointly estimating both the dictionary \((\mD, \vb)\) and the sparse coefficients \(\vz\). When one of these variables is fixed, optimizing over the other becomes relatively easier,\footnote{This observation has motivated alternating minimization methods such as MOD \cite{MOD} and K-SVD \cite{KSVD}.} though still nontrivial in practice. In particular, given a dictionary \(\mD\) and bias \(\vb\), the problem reduces to finding a sparse approximation of \(\vx\), a step commonly referred to as {\it sparse coding}. An efficient method for solving this problem is the proximal gradient method \cite{parikh2014proximal, silva2020efficient}, which is especially suitable when the regularizer \(R(\vz)\) is non-differentiable, such as the $\ell_1$ norm used in Lasso \cite{tibshirani1996regression} or $\ell_0$ norm that directly enforce sparsity \cite{foucart2011hard, 6909888, rajamanoharan2025jumping}.

\paragraph{Proximal gradient methods induce encoders} For a function $r : \R^d \rightarrow \R$, its proximal operator is defined by \cite{parikh2014proximal}
\[
\operatorname{prox}_{r}(\vu) = \argmin_{\vv\in\R^d} \frac{1}{2}\|\vv - \vu\|^2 + r(\vv).
\]
Now starting from an initialization $\vz^{(0)}$, the proximal gradient method for optimizing $\vz$ in \eqref{eq:DL-single-level} performs iterative updates of the form
\begin{equation}
    \vz^{(t+1)} = \operatorname{prox}_{\mu\lambda R} \Big( \vz^{(t)} - \mu \nabla g(\vz^{(t)}) \Big) = \operatorname{prox}_{\mu\lambda R} \Big( \vz^{(t)} - \mu \mD^\top \big( \mD \vz + \vb - \vx \big)) \Big),
\end{equation}
where $\mu>0$ is the step size.
This perspective naturally leads to \emph{unrolled  networks} \cite{gregor2010learning, chen2022learning}, where each proximal gradient step can be interpreted as a layer in a neural network that iteratively refines the latent code $\vz$ while enforcing sparsity \cite{daubechies2004iterative}.
In particular, with $\vz^{(0)} = \vzero$ and $\mu =1$, the first update becomes
\begin{equation}
    \vz^{(1)} = \operatorname{prox}_{\lambda R}\!\left( \mD^\top \vx - \mD^\top \vb \right).
\label{eq:proximal-update}
\end{equation}
Since a single proximal gradient step yields only an approximate solution, inspired by prior work on unrolled networks, we replace the fixed parameters $\mD$ and $\vb$ with learnable counterparts: a trainable weight matrix $\mW$ in place of $\mD$, and a learnable bias vector $\vb_{\text e}$ in place of $-\mD^\top \vb$, thereby yielding a more accurate approximation to the sparse coding solution. Then the update \eqref{eq:proximal-update} becomes
\e
\vz^{(1)} = \operatorname{prox}_{\lambda R}\!\left( \mW^\top \vx + \vb_{\text e} \right),
\label{eq:proximal-solution-sc}\ee
which resembles an encoder.
The following result shows that certain regularizers give rise to proximal operators commonly used in SAEs.
\begin{lemma}
Denote by $\operatorname{ReLU}_\lambda,\operatorname{JumpReLU}_\theta, \operatorname{TopK}_k$ as the following operators:
\e\begin{split}
(\operatorname{ReLU}_{\lambda}(\vu))_i = \max\{u_i - \lambda, 0\}, \quad  (\operatorname{JumpReLU}_\theta(\vu))_i =
    \begin{cases}
        0, & u_i < \theta, \\
        u_i, & u_i \ge \theta,
    \end{cases}, \\(\operatorname{TopK}_k(\vu))_i =
    \begin{cases}
        \max\{u_i,0\}, & i \in \mathcal{T}_k(\vu), \\
        0, & i \notin \mathcal{T}_k(\vu),
    \end{cases}
\end{split}
\label{eq:ReLus}\ee
where $\mathcal{T}_k(\vu)$ denotes the set of indices corresponding to the $k$ largest entries\footnote{In case $k$ largest components are not uniquely defined, one can choose among them—for example, by selecting the components with the smallest indices—to ensure exactly $k$ entries are kept.} of $\vu$. Here $\lambda$, $\theta$ and $k$ are hyper-parameters subject to design choices.

They can be induced by the following choices of sparse regularizers:
\begin{itemize}
[leftmargin=*]
    \item Case I: $R(\vz) = \|\vz\|_1 + \iota_{\{ \vz \ge 0 \}}(\vz)$, then $\operatorname{prox}_{\lambda R} = \operatorname{ReLU}_{\lambda}$;
    \item Case II: $R(\vz) = \|\vz\|_0 + \iota_{\{ \vz \ge 0 \}}(\vz)$, then $\operatorname{prox}_{\lambda R} = \operatorname{JumpReLU}_{\sqrt{2\lambda}}$;
    \item Case III: $R(\vz) = \iota_{\{\|\vz\|_0 \le k, \vz \ge 0\}}(\vz)$, then $\operatorname{prox}_{\lambda R}(\vu) = \operatorname{TopK}_k(\vu)$.
\end{itemize}
Here $\iota_A$ is the indicator function of set $A$, i.e., $\iota_A(\vz) = 0$ if $\vz\in\setA$ and $\iota_A(\vz) = +\infty$ if $\vz\notin \setA$, and $\vz\ge 0$ means $z_i \ge 0$ for all $i$.
\label{lem:proximal-operator}\end{lemma}
A detailed proof is provided in the Appendix \ref{app:proof_lemma}. Note that $\operatorname{ReLU}_\lambda$ reduces to the standard ReLU when $\lambda \rightarrow 0$. The operators $\operatorname{ReLU}\lambda$ and $\operatorname{JumpReLU}_\theta$ are commonly referred to as soft thresholding and hard thresholding (except restricted to the nonnegative orthant), respectively, in signal and image processing, where they are used to enforce sparsity \cite{foucart2011hard, acuna2020soft}.
The TopK operator in \eqref{eq:ReLus} follows the original formulation in \cite{topk}, which includes an additional ReLU to ensure nonnegative activations. Nevertheless, if \(\vu\) has at least $k$ nonnegative entries—which is typically the case since $k$ is much smaller than the ambient dimension $s$—then the ReLU inside TopK is redundant, and the operator simply retains the largest $k$ entries while setting the rest to zero. This phenomenon is also observed in \cite{topk}, where the training curves were found to be indistinguishable. In a nutshell, \Cref{lem:proximal-operator} establishes that several prevalent nonlinearities in SAEs, including ReLU, JumpReLU, and TopK, are precisely the proximal operators of sparse-enforcing regularizers.

\textbf{One-step proximal gradient method leads to Sparse Autoencoders.}
With \Cref{lem:proximal-operator}, applying a one-step proximal gradient method to the sparse coding problem naturally leads to SAEs. Specifically, \eqref{eq:proximal-solution-sc} defines a mapping from an input representation $\vx$ to a sparse code $\vz$, which is then decoded to reconstruct the original representation, formally given by
\begin{equation}
\text{encoder:} \ \vz = \operatorname{prox}_{\lambda R}\!\left( \mathbf{W}^\top \vx + \vb_{\text e} \right), \quad \text{decoder:} \ \wh \vx = \mD\vz + \vb.
\label{eq:DL-only-D}
\end{equation}
Choosing different regularizers $R$ as in \Cref{lem:proximal-operator} yields different variants of SAEs, including the vanilla version with ReLU \cite{cunningham2023sparse}, a version with JumpReLU \cite{rajamanoharan2025jumping}, and one with TopK \cite{topk}. For simplicity, we refer to these as ReLU SAE, JumpReLU SAE, and TopK SAE, respectively. This observation situates diverse SAE architectures within a unified proximal framework, where each activation function is interpreted as the proximal map for a specific regularizer $R$. Consequently, design choices for SAEs correspond directly to the selection of an implicit sparsity-inducing penalty, which in turn provides a principled basis for comparing and extending these models. For instance, our analysis in \Cref{lem:proximal-operator} shows that ReLU SAE corresponds to the $\ell_1$ norm regularizer (a convex relaxation of sparsity) with weight $\lambda \to 0$, whereas JumpReLU and TopK correspond directly to the sparsity-inducing $\ell_0$ norm regularizers with a non-vanishing $\lambda$, thereby enforcing stronger sparsity. This provides a principled explanation for the improved performance of JumpReLU and TopK over ReLU observed in \cite{rajamanoharan2025jumping,topk}.

Substituting \eqref{eq:proximal-solution-sc} into \eqref{eq:DL-single-level} yields the training objective for SAEs \cite{cunningham2023sparse,rajamanoharan2025jumping,topk}:
\begin{equation}
\min_{\substack{\mD, \mW \in \R^{d\times P} \\ \vb\in\R^d, \vb_{\text e} \in \R^P}}
    \E_{\vx}\left[\frac{1}{2}\big\| \vx - {(\bm D\vz + \vb)} \big\|_2^2
    + \lambda R(\vz), \ \text{where}\ \vz = {\operatorname{prox}_{\lambda R}\!\left( \mathbf{W}^\top \vx + \vb_{\text e} \right)}\right].
\end{equation}
In practice, the two instances of $\lambda$ in \eqref{eq:DL-only-D} may be decoupled to provide additional flexibility for hyper-parameter tuning.

The use of a parameterized encoder is a key design choice that circumvents the challenging non-convex optimization in the original dictionary learning formulation \eqref{eq:DL-single-level}, which requires simultaneous optimization over the sparse codes $\vz$ and the dictionary parameters $\mD$ and $\vb$. By decoupling this joint optimization, SAEs yield a more tractable training procedure. The encoder arises as a single proximal gradient step, augmented with uncoupled, learnable parameters for the dictionary and bias to reduce the approximation error relative to exact sparse coding. Consequently, the training problem \eqref{eq:DL-only-D} can be efficiently solved via stochastic gradient descent, and SAEs can be implemented efficiently at inference time, making them attractive for interpretability research.

This perspective also provides a principled foundation for developing SAE variants with improved performance. For example, by incurring additional computational cost, one may extend \eqref{eq:proximal-solution-sc} to multi-step variants, yielding multi-layer encoders \cite{tolooshams2022stable} that produce more accurate sparse codes and potentially capture finer-grained structure in the representation space. We leave this direction to future work. In the next subsection, we turn to SAEs induced by alternative sparsity regularizers.

\subsection{Beyond non-negativity: Sparse Autoencoders with AbsTopK}

The proximal perspective developed above suggests that design choices for SAEs can be interpreted as the selection of the sparsity-inducing penalty. While this view explains their sparsity-inducing effect, it also reveals a fundamental limitation of current SAEs in \Cref{eq:ReLus}: they prompt sparsity but also enforce non-negativity, discarding half of the representation space. As many semantic axes are naturally bidirectional (e.g., \emph{male v.s. female}, \emph{positive v.s. negative sentiment}), restricting sparse codes to be nonnegative fragments these concepts into two separate directions or collapses one side entirely.

\paragraph{Fragmentation of Conventional SAE}
To formalize this, consider a single semantic concept direction $\vh$ in \eqref{linear_combination} represented by a dictionary atom $\vd \in \R^d$. An ideal sparse code would represent concepts along this axis as $\alpha \vd$, where the sign of the scalar $\alpha$ encodes directionality. However, under the non-negativity constraint $\vz \geq 0$, this is impossible. Instead, a standard SAE must allocate two distinct dictionary atoms, $\vd_i$ and $\vd_j$, oriented in opposite directions, with nonnegative activations $z_i \geq 0$ and $z_j \geq 0$ respectively. Each atom is activated only for one direction, leading to a fragmented representation that arises directly from the non-negativity constraint. This fragmentation is a direct consequence of the non-negativity constraint.

\paragraph{Removing non-negativeness as a remedy.}  To address this issue, we propose using a sparse regularizer without the non-negativity constraint. Different variants of sparse regularizers can be considered, with representative examples discussed in \Cref{lem:proximal-operator}. In this work, we adopt the $\ell_0$ norm due to its simplicity and its direct connection to sparsity. Specifically, in the dictionary learning formulation \eqref{eq:DL-single-level}, we use the regularizer $R(\vz) = \iota_{\{\|\vz\|_0 \le k\}}$ which removes the non-negativity constraint present in the TopK-inducing regularizer. The corresponding proximal operator is
\e
\operatorname{prox}_{\lambda R}(\vu)
= \arg\min_{\vz \in \mathbb R^d}\; \tfrac{1}{2}\|\vu-\vz\|_2^2
\quad \text{s.t.}\quad \|\vz\|_0 \le k,
\ee
whose closed-form solution is further given by
\begin{equation}
\bigl(\operatorname{prox}_{R}(\vu)\bigr)_i = (\operatorname{AbsTopK}_k(\vu))_i =
\begin{cases}
u_i, & i \in {\mathcal{H}}_k(\vu),\\[2pt]
0,   & i \notin {\mathcal{H}}_k(\vu),
\end{cases}
\end{equation}
where $\calH_k$ denotes the indices of the $k$ largest (in modulus) components\footnote{Similarly, if the $k$ largest components are not uniquely defined, one can, for instance, select those with the smallest indices to ensure exactly $k$ entries are retained.}. In words, this operator preserves the $k$ largest-magnitude components of a vector and sets all others to zero. In the compressive sensing literature, it is referred to as the {\it hard thresholding operator} \cite{foucart2011hard}. Here, we refer to it as Absolute TopK (AbsTopK) to distinguish it from the TopK operator commonly used in SAE.

This principle of hard thresholding can also be applied to JumpReLU, introducing a threshold on both positive and negative activations. This achieves a similar effect by eliminating small-magnitude features and enforcing sparsity. However, to isolate and directly test our core hypothesis, the value of representing concepts along a bipolar axis, this work focuses on AbsTopK, as it provides the most direct implementation of a global $k$-sparsity constraint. We remain JumpReLU variants for future investigation.

\paragraph{AbsTopK SAE.}
Following the derivation in the previous section, we integrate the AbsTopK nonlinearity operator into the framework \eqref{eq:DL-only-D} to obtain a new SAE architecture, which we term AbsTopK SAE:
\begin{equation}
\begin{aligned}
\vz = \operatorname{AbsTopK}(\mW^\top \vx + \vb_{\text e}), \ \wh{\vx} = \mD \vz + \vb.
\end{aligned}
\end{equation}
The overall training problem becomes
\begin{equation}
\min_{\substack{\mD, \mW \in \R^{d\times P} \\ \vb\in\R^d, \vb_{\text e} \in \R^P}}
    \E_{\vx}\left[\frac{1}{2}\big\| \vx - {(\bm D\vz + \vb)} \big\|_2^2, \ \text{where}\ \vz = \operatorname{AbsTopK}(\mW^\top \vx + \vb_{\text e})\right].
\end{equation}
By design, AbsTopK preserves both positive and negative activations, enabling a single feature to capture contrastive concepts along a unified semantic axis. This simple modification circumvents the fragmentation induced by non-negativity constraints, and yields features that more faithfully reflect the bidirectional structure of semantic representations.

\section{Experiments: Empirical Validation of SAE behavior}

To empirically validate our theoretical claims and demonstrate the practical advantages of the AbsTopK operator, we perform a suite of experiments which involve training JumpReLU, TopK, and AbsTopK SAEs on \texttt{monology/pile-uncopyrighted} \cite{pile} across the GPT2-SMALL, Pythia-70M, Gemma2-2B, and Qwen3-4B models \cite{radford2019language, biderman2023pythia, gemma_2024, qwen3}.
To compare the different SAEs, we evaluate their performance along several dimensions: (i) reconstruction quality on base datasets, (ii) effectiveness on a range of steering tasks, and (iii) impact on general capabilities of the models. For further experimental details and extended results, we refer the reader to Appendix~\ref{appendix:experimental-details}.

\begin{figure*}[ht]
\includegraphics[width=\linewidth]{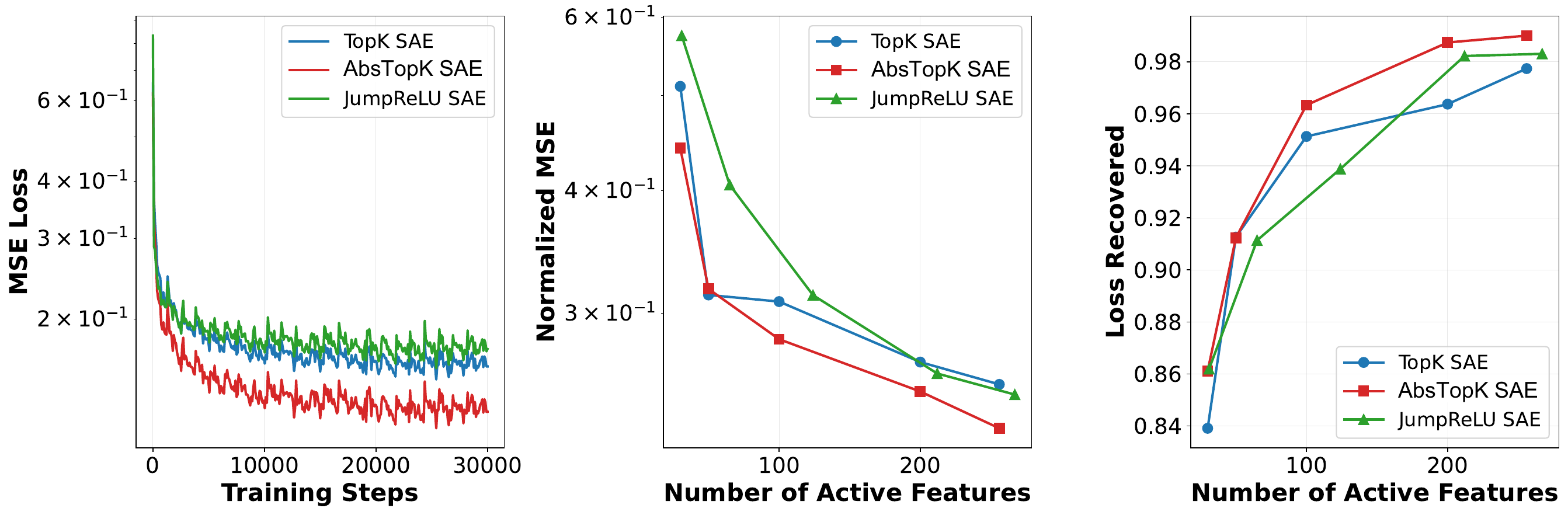}
\caption{
\textbf{Performance comparison of JumpReLU, TopK, and AbsTopK SAEs on Qwen3 4B Layer 20}, showing \textbf{(a)} MSE Training Loss, \textbf{(b)} Normalized MSE, and \textbf{(c)} Loss Recovered. Additional results across models and layers are provided in Appendix \ref{app:core_metrics}.
}
\label{fig:absK_results}
\end{figure*}

\begin{figure}[!ht]
    \centering
        \includegraphics[width=0.8\linewidth]{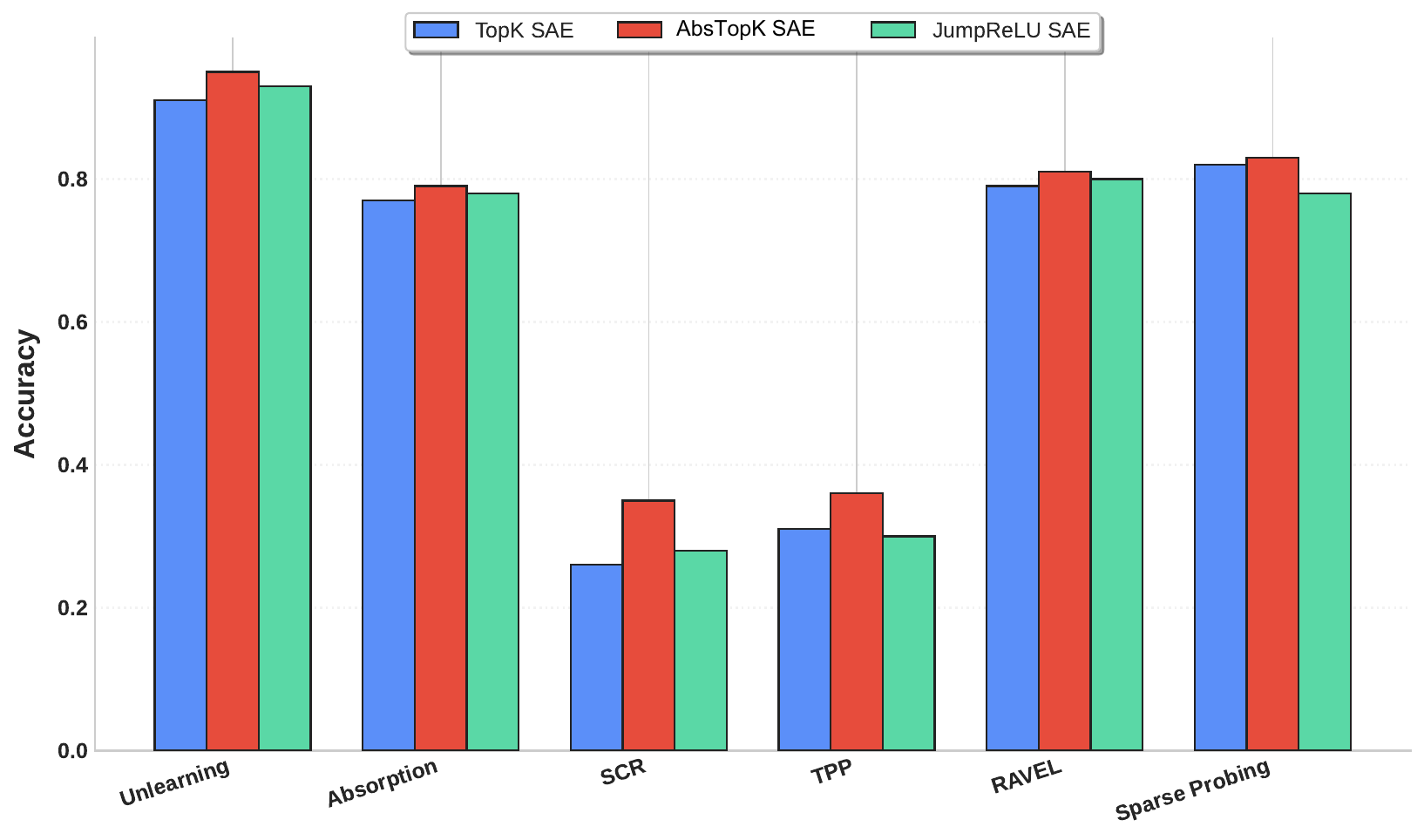}
    \caption{\textbf{Performance comparison of SAE variants (TopK, AbsTopK, and JumpReLU) across tasks on Qwen3-4B Layer 18.} For all tasks, higher scores indicate better performance; the Unlearning and Absorption scores have been transformed as $1-$original score to maintain this consistency. For more details, see Appendix~\ref{appendix:sae_task_details}.}
    \label{fig:sae_task_comparison}
\end{figure}
\vspace{-0.8em}
\subsection{Unsupervised Metrics}

This section presents a comparative evaluation of SAE architectures, utilizing a suite of complementary metrics engineered to assess distinct facets of model performance. The investigation encompasses three primary analyses: (a) an examination of the training mean squared error (MSE) to evaluate optimization stability and convergence rates; (b) the measurement of normalized reconstruction error as a function of feature sparsity to ascertain representational fidelity; and (c) a relative cross-entropy loss recovered score to determine the preservation of language modeling performance. For Topk and AbsTopK, sparsity is explicitly controlled by directly specifying the number of active features $k$; in contrast, for JumpReLU, sparsity is varied by manually adjusting the threshold parameter $\theta$, thereby simulating different sparsity levels.

The normalized reconstruction error in (b) is defined as $\text{nMSE}(\vx, \hat{\vx}) = \|\vx - \hat{\vx}\|_2^2 / \|\vx\|_2^2$ \cite{topk}, thereby controlling for scale differences across representations. The Loss Recovered score in (c) measures how well SAE reconstructions preserve predictive performance \cite{karvonen2025saebench}, defined as $(H^* - H_0) / (H_{\text{orig}} - H_0)$, where $H_{\text{orig}}$ is the cross-entropy of the original model, $H^*$ that after substitution, and $H_0$ under zero-ablation, with values closer to one indicating better preservation.

The empirical results of this evaluation delineate a distinct performance hierarchy among the three architectures under consideration. The AbsTopK architecture demonstrates the most favorable performance characteristics. Specifically, the AbsTopK model consistently yields lower reconstruction errors across most tested sparsity levels and manifests minimal cross-entropy degradation. This demonstrates its superior capacity for preserving the integrity of the language model's performance under conditions of representational compression.

The observed disparities in performance can be attributed to the relative expressiveness of the constraints inherent to each architectural formulation. The TopK and JumpReLU architectures both impose a non-negativity constraint on activations. This imposition results in a conical decomposition of the feature space, which has the effect of fragmenting concepts that are inherently bidirectional in nature. In contrast, the AbsTopK architecture accommodates both positive and negative activation values, a flexibility that facilitates a more faithful linear decomposition of the model's latent states. This capacity for bidirectionality furnishes a more compact and interpretable representational basis, as it permits a single feature to encode oppositional concepts through its algebraic sign. As will be further substantiated in subsequent qualitative analyses, this fundamental property enables the AbsTopK model to acquire dictionary atoms that exhibit a closer alignment with the underlying conceptual structures embedded within the model's representations.

\subsection{Results on Probe and Steering Tasks}

To assess the utility of learned SAE features for model control, a comprehensive benchmarking evaluation was conducted across a diverse suite of steering and probing tasks. These tasks were specifically designed to probe various dimensions of feature quality, from basic concept representation to the capacity for precise interventional control. A detailed methodological overview for each metric is provided in the Appendix \ref{appendix:sae_task_details}.

The empirical results, as shown in Table~\ref{fig:sae_task_comparison}, demonstrate the superiority of the AbsTopK methodology. Across the entire suite of evaluated tasks, AbsTopK SAE outperforms both the TopK SAE and JumpReLU SAE baselines. This performance advantage is especially conspicuous in bidirectional steering metrics, such as SCR, which directly quantify the reliability of interventions. In these critical evaluations, AbsTopK shows marked improvements over the alternatives.

We posit that this consistent outperformance is directly attributable to the core mechanism of the AbsTopK methodology: the retention of both positive and negative feature activations. Unlike TopK approaches, which enforce a hard sparsity constraint that discards all but the most prominent positive activations, AbsTopK preserves a richer, more complete semantic representation. This retention is critical for interventions that require nuanced and bidirectional control. By encoding not only the presence of a concept but also its negation or semantic opposition, AbsTopK features provide a more robust and granular basis for manipulation.

\subsection{Empirical Results on Steering vs.\ Utility}

\begin{table}[t]
\centering
\caption{\textbf{Performance comparison on MMLU ($\uparrow$) and HarmBench ($\uparrow$) across steering methods.} The best result among all methods for each metric is highlighted in \textbf{bold}. For details of the steering methods, see Appendix~\ref{app:steer_methods}.}
\resizebox{\linewidth}{!}{
\begin{tabular}{lllccccc}
\toprule
\textbf{Model} & \textbf{Layer} & \textbf{Metric} & \textbf{Original} & \textbf{JumpReLU SAE} & \textbf{TopK SAE} & \textbf{AbsTopK SAE} & \textbf{DiM} \\
\midrule
\multirow{4}{*}{Qwen3-4B}
 & \multirow{2}{*}{18} & MMLU & 77.3 & 75.0 & 75.2 & \textbf{75.9} & 75.8 \\
 &                       & HarmBench & 17.0 & 79.1 & 78.2 & \textbf{81.3} & 80.6 \\
 & \multirow{2}{*}{20} & MMLU & 77.3 & 75.7 & 75.0 & \textbf{76.4} & \textbf{76.4} \\
 &                       & HarmBench & 17.0 & 78.5 & 77.0 & 79.0 & \textbf{80.0} \\
\midrule
\multirow{4}{*}{Gemma2-2B}
 & \multirow{2}{*}{12} & MMLU & 52.2 & 48.8 & 49.1 & \textbf{51.3} & 51.0 \\
 &                       & HarmBench & 19.0 & 69.5 & 69.8 & 70.2 & \textbf{70.8} \\
 & \multirow{2}{*}{16} & MMLU & 52.2 & 48.2 & 48.5 & \textbf{51.0} & 50.8 \\
 &                       & HarmBench & 19.0 & 69.8 & 70.2 & 71.7 & \textbf{72.0} \\
\bottomrule
\end{tabular}}
\label{tab:safety-utility-tradeoff}
\end{table}

Model steering confronts a fundamental tradeoff: enhancing specific behaviors often degrades general capabilities. It has often been assumed in prior literature that DiM interventions are more effective for specific concept manipulation than SAEs despite their reliance on labeled data and limitation to extracting only a single concept vector \cite{arditi2024refusal, wu2025axbench, zhu2025emergence}. To systematically evaluate this trade-off, we conducted an empirical study measuring general capability preservation via the MMLU benchmark \cite{hendryckstest2021} and safety alignment using HarmBench \cite{mazeika2024harmbench}. For this evaluation, we focus on Qwen and Gemma models, as smaller models, Pythia-70M and GPT-2 Small, only have very low score on MMLU benchmark.

As shown in Table~\ref{tab:safety-utility-tradeoff}, the empirical results indicate that conventional SAE steering methods successfully improve safety metrics but at a detriment to general performance. In contrast, the proposed AbsTopK methodology achieves a more optimal balance between these competing objectives. It facilitates substantial enhancements in safety alignment on HarmBench while simultaneously mitigating the degradation of MMLU scores. Compared to DiM, AbsTopK is competitive on safety, sometimes slightly lower, but consistently retains more general ability. This pattern highlights that carefully designed SAE steering can rival and, in some cases, surpass intervention strategies that rely on labeled data.

\section{Conclusion}

This work identifies the non-negativity constraint in SAEs as a core cause of semantic feature fragmentation. In response, we introduce the AbsTopK operator, which replaces this constraint with direct k-sparsity enforced via an $\ell_0$ proximal operator. This modification enables single features to capture bipolar semantics, and our empirical results confirm that AbsTopK yields reconstructions of superior compactness and fidelity. Our work pioneers a shift towards bipolar sparse representations and suggests future research into more efficient, neurally-plausible approximations of the $\ell_0$ operator for large-scale models.

\bibliographystyle{ieeetr}
\bibliography{learning}

\appendix
\section{Related Works}

\paragraph{Sparse Dictionary Learning}
The dictionary learning problem \eqref{eq:DL-single-level} is highly nonconvex. Over the past decades, numerous heuristic methods have been proposed to solve it efficiently \cite{MOD,KSVD,tovsic2011dictionary}.
Significant effort has also been devoted to addressing the nonconvexity and establishing theoretical guarantees for exact recovery, including approaches based on convex relaxation  \cite{spielman2012exact}, semidefinite programming \cite{barak2015dictionary}, landscape analysis of simplified formulations that estimate one atom at a time
\cite{sun2016complete,gilboa2019efficient,zhu2019linearly,qu2020geometric}, alternating minimization \cite{arora2015simple,Chatterji2017AlternatingMF},  as well as global convergence guarantees to critical points \cite{ 6909888, 10.5555/2834535, 7293682}. Subsequent work has explored unrolled methods \cite{malzieux2022understanding,tolooshams2022stable,chen2022learning}, analyzed convergence for solving the sparse coding problem under the assumption of a fixed dictionary
\cite{gregor2010learning, Tang2020DeepTA, massoli2024variational}, and examined gradient stability \cite{tolooshams2022stable, malzieux2022understanding}.

Building on these works, we re-examine recently popularized SAEs for interpretability through the lens of unrolled dictionary learning. This perspective reveals a direct correspondence between activation functions and the proximal mappings of sparse regularizers, thereby situating SAEs within the broader framework of dictionary learning. Leveraging this connection, we introduce a new activation, \emph{AbsTopK}, which removes the non-negativity constraint imposed in prior designs and enables a single dictionary feature to encode bidirectional semantic axes. Our principal contribution is thus a formal alignment of SAE architecture with the intrinsic geometry of semantic representation—distinct from classical theories focused on signal recovery.

\paragraph{Mechanistic interpretability}

Sparse autoencoders (SAEs) have emerged as a central tool in mechanistic interpretability, serving as a dictionary learning approach for concept-level explainability \cite{kim2018interpretability}.
A variety of architectures have been proposed, including ReLU SAEs \cite{bricken2023monosemanticity}, TopK SAEs \cite{topk}, JumpReLU SAEs \cite{rajamanoharan2025jumping}, gated SAEs \cite{rajamanoharan2024improving}, Batch TopK SAEs \cite{bussmann2024batchtopk}, and ProLU SAEs \cite{neill2025compute}, among others.
Indeed, these models have demonstrated a remarkable ability to capture a diverse spectrum of interpretable concepts within their latent representations, ranging from abstract notions like refusal, gender, and writing script \cite{bricken2023monosemanticity,templeton2024scaling, hegde2024effectiveness}, to visual elements such as foreground/background separation \cite{thasarathan2025universal}, and even the fundamental structures of proteins \cite{simon2024interplm}.

Despite these successes, a growing body of work has highlighted limitations of the SAE paradigm.
Simple prompting baselines have been shown to outperform SAE interventions in model control \cite{wu2025axbench, bhalla2025towards}, with similar conclusions reported in formal language settings \cite{menon2024analyzing}.
Other critiques question the linear representation hypothesis underlying classical SAE design, showing that features can be multidimensional or nonlinear \cite{engels2024decomposing, engels2025not, peng2025use, wang2025beyond}.
Moreover, multiple studies demonstrate severe algorithmic instability: two SAEs trained on the same data but with different random seeds may yield divergent feature dictionaries, leading to inconsistent interpretations \cite{ayonrinde2024interpretability, SAEsAreHighlyDatasetDependent, colin2025local}.
These observations suggest that, while SAEs provide a promising path toward interpretability, their current formulation suffers from fragility and non-canonical representations.

Our work responds to this need by placing SAE nonlinearities in the broader dictionary learning framework, using a proximal perspective.
This unifying view clarifies the connection between popular activation functions and sparse regularizers and motivates our proposed \emph{AbsTopK}, which enables bidirectional semantic representation.
\section{Experimental Setup}
\label{appendix:experimental-details}

In this appendix, we describe the architecture and training setup of our SAEs.
For all experiments, we trained on the \texttt{monology/pile-uncopyrighted} \cite{pile} dataset.

Architecturally, the SAEs are comprised of a single, overcomplete hidden layer which incorporates a sparsifying nonlinearity. The encoder component projects residual activations into a latent space of higher dimensionality, while the decoder component reconstructs the original residual dimension from these latent representations. A fixed expansion factor of 16 was uniformly applied across all models.

For comparative analysis, three distinct variants of the SAE were trained: TopK, AbsTopK, and JumpReLU. In the TopK and AbsTopK configurations, exact k-sparsity was enforced upon the latent representation, with the sparsity hyperparameter, k, and the specific layers targeted for intervention being systematically selected for each foundational model:
\begin{itemize}
    \item \texttt{EleutherAI/pythia-70m} \cite{biderman2023pythia}: $k = 51$, layers: 3, 4.
    \item \texttt{google/gemma-2-2b} \cite{gemma_2024}: $k = 230$, layers: 12, 16.
    \item \texttt{Qwen/Qwen3-4B} \cite{qwen3}: $k = 256$, layers: 18, 20.
    \item \texttt{openai-community/gpt2} \cite{radford2019language}: $k = 76$, layers: 6, 8.
\end{itemize}
Here, $k$ was set to approximately one-tenth of the hidden dimension for each model, and the intervention layers were selected from the middle of the network to capture representative latent features \cite{arditi2024refusal}. In contrast, the JumpReLU models adopted the same configuration as in prior work \cite{rajamanoharan2025jumping, bussmann2024batchtopk}.

The optimization for all models was performed using the Adam algorithm over a duration of 30,000 training steps, with a consistent batch size of 4096. A learning rate of 3e-4 was configured, complemented by Adam's momentum parameters, $\beta_1 = 0.9$, and $\beta_2 = 0.99$. And we used a bandwidth parameter of $0.001$ across all experiments.
\section{Proof of Lemma1}
\label{app:proof_lemma}
\begin{proof}
We prove the result by deriving the proximal operator corresponding to each regularizer separately.
\paragraph{Case I: ReLU.}
Note that $R(\vz)$ is separable as
\[
R(\vz)=\|\vz\|_1 + \iota_{\{\vz \ge 0\}}(\vz) = \sum_i \left(  |z_i| + \iota_{\{z_i \ge 0\}}(z_i) \right) ,
\]
which implies that the proximal operator is also separable, i.e., $(\operatorname{prox}_{\lambda R}(\vu))_i$ is equivalent to the following scalar proximal problem

\begin{align*}
\operatorname{prox}_{\lambda R}(u) & =\arg\min_{z\in \R}\; \frac{1}{2}(z-u)^2 + \lambda |z| + \iota_{\{z\ge 0\}}(z)\\
& = \arg\min_{z\ge 0}\; \frac{1}{2}(z-u)^2 + \lambda z\\
& = \max\{u - \lambda,\,0\}.
\end{align*}
Therefore, the proximal operator induces the ReLU operator, with a shift by $\lambda$:
\[
\boxed{\;(\operatorname{prox}_{\lambda R}(\vu)) \;=\; \max\{\vu - \lambda,\,0\},\;}
\]
which reduces to the standard ReLU when $\lambda \to 0$. In this case, however, the operator no longer encourages sparsity. When $\lambda > 0$, the effect is equivalent to introducing a bias term that suppresses small activations and thereby promotes sparsity. In practice, this restriction can be relaxed: during training, gradient descent can learn a separate bias parameter for each entry.

\paragraph{Case II: JumpReLU.} Similarly, $R(\vz)$ is also separable as
\[
R(\vz)=\|\vz\|_0 + \iota_{\vz \ge 0}(\vz) =
\sum_i \left(
   \mathbf{1}(z_i \neq 0) + \iota_{\{z_i \ge 0\}}(z_i)
\right).
\]
where $\mathbf{1}(z_i \neq 0) = \begin{cases} 1,  z_i \neq 0,\\ 0, z_i = 0. \end{cases}$ Thus, it suffices to first consider the following scalar proximal operator
\begin{align*}
\operatorname{prox}_{\lambda R}(u) & =\arg\min_{z\in\mathbb R}\; \frac{1}{2}(z-u)^2 + \lambda \mathbf{1}(z \neq 0) + \iota_{\{z\ge 0\}}(z)\\
& = \arg\min_{z\ge 0}\; \underbrace{\frac{1}{2}(z-u)^2 + \lambda \mathbf{1}(z \neq 0)}_{\xi(z)}.
\end{align*}
Note that within the region $z\ge 0$, $\xi$ achieve its minimum at either 0 or $u$. Setting $\xi(u) = \lambda = \xi(0) = \frac{1}{2}u^2$ yields $u = \sqrt{2\lambda}$. One can verify that $\xi$ achives its minimum at $u$ when $u\ge \sqrt{2\lambda}$, and at $0$ otherwise. Hence, the proximal operator induces the JumReLU with parameter $\sqrt{2\lambda}$:
\[
\boxed{\;(\operatorname{prox}_{\lambda R}(\vu))_i \;=\; \begin{cases}
u, & u \ge \sqrt{2\lambda},\\[6pt]
0, & u < \sqrt{2\lambda}.
\end{cases}\;}
\]

\paragraph{Case III: TopK.}

For this case, the corresponding proximal operator reduces to a Euclidean projection onto the
feasible set:
\e
\operatorname{prox}_{\lambda R}(\vu)
= \arg\min_{\vz \in \mathbb R^d}\; \tfrac{1}{2}\|\vu-\vz\|_2^2
\quad \text{s.t.}\quad \|\vz\|_0 \le k,\ \vz \ge 0.
\ee

Given the quadratic objective and the non-negativity constraint, the optimal choice on any candidate support $S$ with $|S|\le k$ is
\e
z_i =
\begin{cases}
\max\{u_i,0\}, & i\in S,\\
0, & i\notin S.
\end{cases}
\ee
Thus, the minimization problem reduces to selecting the index set $S$ that captures the $k$ largest nonnegative entries of $\vu$. Formally, letting $\mathcal{T}_k(\vz)$ denote the set of indices corresponding to the $k$ largest entries of $\vz$, the proximal operator becomes
\e
\boxed{\;[\operatorname{prox}_{\lambda R}(\vu)]_i =
\begin{cases}
\max\{u_i,0\}, & i\in \calT_k(\vz),\\[3pt]
0, & i\notin \calT_k(\vz).
\end{cases}\;}
\ee
\end{proof}

\section{Unsupervised Metrics on All Models} \label{app:core_metrics}

\begin{figure}[!ht]
    \centering
    \includegraphics[width=0.8\linewidth]{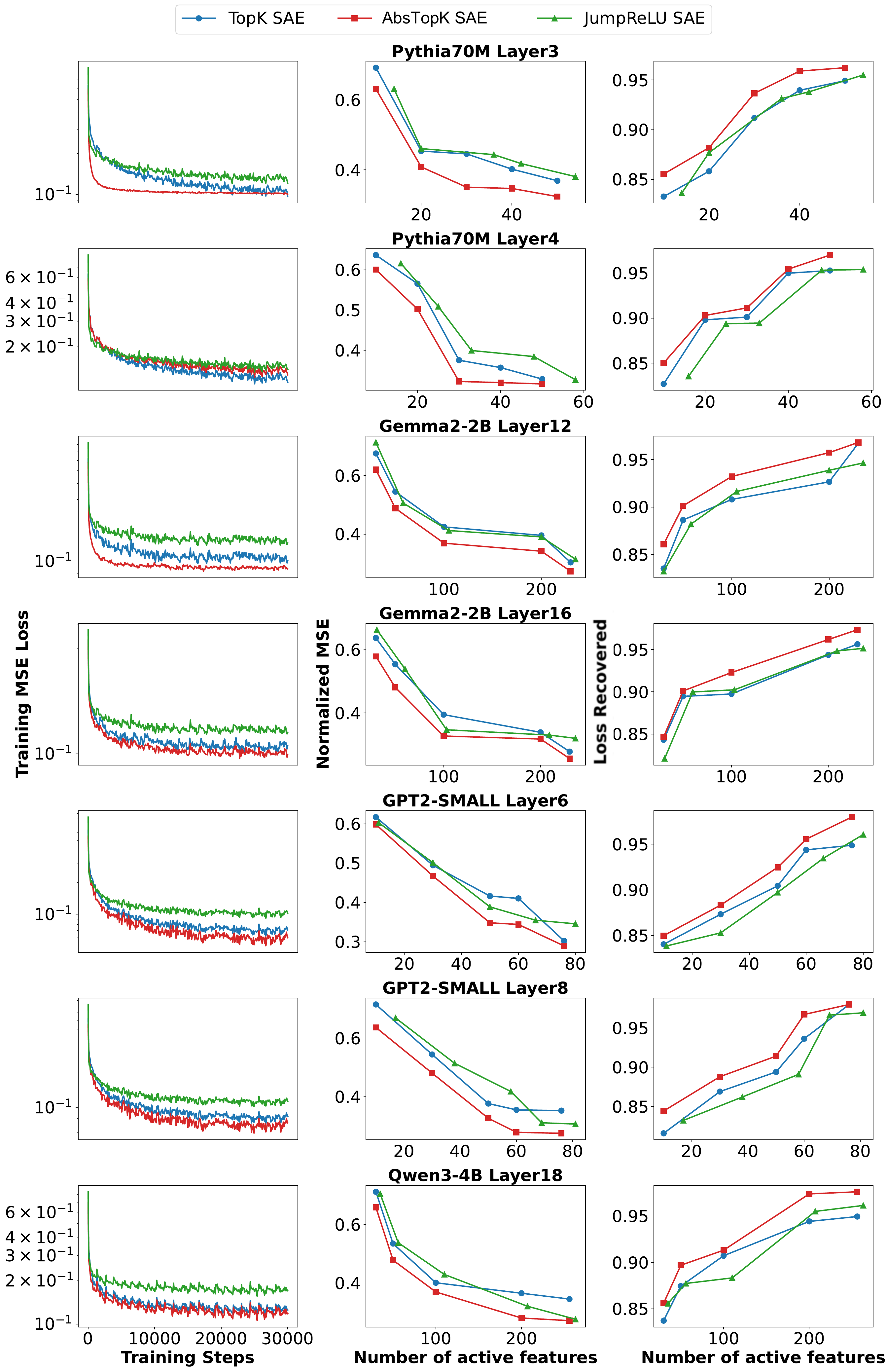}
\caption{
\textbf{Performance comparison of JumpReLU, TopK, and AbsTopK SAEs on all other models and layers}, showing \textbf{(a)} MSE Training Loss, \textbf{(b)} Normalized MSE, and \textbf{(c)} Loss Recovered.
}\label{fig:core_all}
\end{figure}

This section presents the unsupervised metrics from our model evaluations. We tested each model with a specific set of k values. For the Pythia model, we used k-values of 10, 20, 30, 40, and 50. The evaluation of the Gemma model involved k values of 30, 50, 100, 200, and 230. For the GPT model, the k values were 10, 30, 50, 60, and 76. Lastly, the Qwen model was tested with k values of 30, 50, 100, 200, and 256.

As shown in Figure~\ref{fig:core_all}, across the majority of evaluated models, we observe that AbsTopK achieves lower training MSE, reduced normalized reconstruction error, and better preservation of language modeling performance relative to both TopK and JumpReLU. This consistent advantage across these metrics provides evidence for the effectiveness and robustness of the AbsTopK method.. In particular, while TopK and JumpReLU sometimes exhibit competitive performance in isolated settings, AbsTopK maintains robustness across architectures and layers, thereby demonstrating the superiority of our proposed formulation.

\section{Steering and Probe Task on All Models}
\label{appendix:sae_task_details}

\begin{table}[!ht]
\centering
\caption{\textbf{Performance comparison of SAE variants across tasks on all other models and layers.} For all tasks, higher scores indicate better performance; the Unlearning and Absorption scores have been transformed as $1-$original score to maintain this consistency.}
\label{tab:task-performance}
\resizebox{\linewidth}{!}{
\begin{tabular}{llcccccc}
\toprule
\textbf{Model} & \textbf{Method} & Unlearning & Absorption & SCR & TPP & RAVEL & Sparse Probing \\
\midrule
\multirow{3}{*}{Gemma2-2B L12}
  & AbsTopK & 0.93 & 0.73 & 0.27 & 0.34 & 0.73 & 0.76 \\
  & TopK    & 0.88 & 0.76 & 0.20 & 0.29 & 0.70 & 0.71 \\
  & JumpReLU& 0.90 & 0.75 & 0.22 & 0.30 & 0.71 & 0.73 \\
\midrule
\multirow{3}{*}{Gemma2-2B L14}
  & AbsTopK & 0.91 & 0.70 & 0.27 & 0.42 & 0.71 & 0.70 \\
  & TopK    & 0.89 & 0.68 & 0.21 & 0.36 & 0.74 & 0.67 \\
  & JumpReLU& 0.94 & 0.69 & 0.23 & 0.39 & 0.72 & 0.69 \\
\midrule
\multirow{3}{*}{Pythia-70M L3}
  & AbsTopK & 0.75 & 0.54 & 0.20 & 0.22 & 0.64 & 0.66 \\
  & TopK    & 0.71 & 0.47 & 0.15 & 0.14 & 0.62 & 0.60 \\
  & JumpReLU& 0.73 & 0.50 & 0.17 & 0.21 & 0.63 & 0.61 \\
\midrule
\multirow{3}{*}{Pythia-70M L4}
  & AbsTopK & 0.79 & 0.53 & 0.21 & 0.23 & 0.68 & 0.57 \\
  & TopK    & 0.72 & 0.50 & 0.16 & 0.21 & 0.69 & 0.61 \\
  & JumpReLU& 0.77 & 0.51 & 0.17 & 0.20 & 0.61 & 0.62 \\
\midrule
\multirow{3}{*}{GPT2-small L6}
  & AbsTopK & 0.74 & 0.66 & 0.18 & 0.22 & 0.60 & 0.54 \\
  & TopK    & 0.80 & 0.63 & 0.14 & 0.19 & 0.57 & 0.50 \\
  & JumpReLU& 0.77 & 0.65 & 0.15 & 0.20 & 0.58 & 0.52 \\
\midrule
\multirow{3}{*}{GPT2-small L8}
  & AbsTopK & 0.75 & 0.67 & 0.23 & 0.28 & 0.51 & 0.59 \\
  & TopK    & 0.71 & 0.67 & 0.15 & 0.20 & 0.48 & 0.55 \\
  & JumpReLU& 0.73 & 0.67 & 0.18 & 0.23 & 0.49 & 0.57 \\
\midrule
\multirow{3}{*}{Qwen3-4B L18}
  & AbsTopK & 0.95 & 0.79 & 0.35 & 0.36 & 0.81 & 0.83 \\
  & TopK    & 0.91 & 0.77 & 0.26 & 0.31 & 0.79 & 0.82 \\
  & JumpReLU& 0.93 & 0.78 & 0.28 & 0.30 & 0.80 & 0.78 \\
\midrule
\multirow{3}{*}{Qwen3-4B L20}
  & AbsTopK & 0.95 & 0.80 & 0.32 & 0.45 & 0.85 & 0.81 \\
  & TopK    & 0.92 & 0.77 & 0.27 & 0.36 & 0.76 & 0.84 \\
  & JumpReLU& 0.93 & 0.78 & 0.29 & 0.39 & 0.81 & 0.83 \\
\bottomrule
\end{tabular}
}
\end{table}

\subsection{Task Description}
We provide an overview of the tasks employed in the SAEBench evaluation for SAEs. We do not utilize the Automated Interpretability (AutoInterp) evaluation, as its reliability has been questioned \cite{heap2025sparse}. For detalied methodology, we refer readers to the original SAEBench paper~\cite{karvonen2025saebench}.

\subsubsection{Feature Absorption}

Sparsity incentives can cause a SAE to engage in feature absorption, a phenomenon where correlated features are merged into a single latent representation. This process arises when a direct implication exists between two concepts, such that concept $A$ always implies concept $B$. To reduce the number of active latents, the SAE might absorb the feature for $A$ into the latent for $B$. For example, a feature for "starts with S" could be absorbed into a more general latent for "short." While this merging improves computational efficiency, it compromises interpretability by creating gerrymandered features that represent multiple, distinct concepts.

To quantify feature absorption, we employ a first-letter classification task, following the methodology of previous studies \cite{chanin2025a}. First, a supervised logistic regression probe is trained on tokens containing only English letters to establish ground-truth feature directions. Next, K-sparse probing is applied to the SAE's latents to identify the primary latent corresponding to each feature, using a threshold of $\tau_\text{fs}=0.03$ to account for potential feature splits. For test set tokens where main latents fail but the probe succeeds, additional SAE latents are included if they satisfy cosine similarity with the probe of at least $\tau_\text{ps}=0.025$ and a projection fraction of at least $\tau_\text{pa}=0.4$. All parameter values are chosen following the original SAEBench settings \cite{karvonen2025saebench}. To make the results more interpretable and such that higher values indicate stronger unlearning, we present the final scores as $1 - \text{original value}$.

\subsubsection{Unlearning}
SAEs are evaluated on their ability to selectively remove knowledge while maintaining performance on unrelated tasks \cite{farrell2025applying}. We use the WMDP-bio dataset \cite{li2024wmdp} for unlearning and MMLU \cite{hendryckstest2021} to assess general abilities.

The intervention methodology involves clamping selected WMDP-bio SAE feature activations to negative values whenever the corresponding features activate during inference. To evaluate broader model effects, we also measure performance on the MMLU benchmark \cite{hendryckstest2021}. The final evaluation reports the highest unlearning effectiveness on WMDP-bio while ensuring MMLU accuracy remains above 0.99, thereby quantifying optimal unlearning performance under constrained side effects. To make the results more interpretable and such that higher values indicate stronger unlearning, we present the final scores as $1 - \text{original value}$.

\subsubsection{Spurious Correlation Removal (SCR)}
SCR \cite{karvonen2024evaluating} evaluates the ability of SAEs to disentangle latents corresponding to distinct concepts. We conduct experiments on datasets known for spurious correlations, such as Bias in Bios \cite{de2019bias} and Amazon Reviews \cite{hou2024bridging}, which contain two binary gender labels. For each dataset, we create a balanced set containing all combinations of profession (professor/nurse) and gender (male/female), as well as a biased set including only male+professor and female+nurse combinations. A biased classifier $C$ is first trained on the biased set and then debiased by ablating selected SAE latents.

We quantify SCR using the normalized evaluation score:
\begin{equation}
S_\text{SHIFT} = \frac{A_\text{abl} - A_\text{base}}{A_\text{oracle} - A_\text{base}},
\end{equation}
where $A_\text{abl}$ is the probe accuracy after SAE feature ablation, $A_\text{base}$ is the baseline accuracy before ablation, and $A_\text{oracle}$ is the skyline accuracy obtained by a probe trained directly on the desired concept. Higher $S_\text{SHIFT}$ values indicate more effective removal of spurious correlations. This score represents the proportion of improvement achieved through ablation relative to the maximum possible improvement, enabling fair comparison across classes and models.

\subsubsection{Targeted Probe Perturbation (TPP)}

TPP \cite{marks2025sparse} extends the SHIFT methodology to multiclass natural language processing datasets. For each class $c_i$ in a dataset, we select the most relevant SAE latents $L_i$. We then evaluate the causal effect of ablating $L_i$ on linear probes $C_j$ trained to classify each class $c_j$.

Let $A_j$ denote the accuracy of probe $C_j$ before ablation, and $A_{j \setminus i}$ the accuracy after ablating $L_i$. We define the accuracy change as
\e
\Delta A_{j \setminus i} = A_{j \setminus i} - A_j.
\ee
The TPP score is then
\e
S_\text{TPP} = \mathbb{E}_{i=j}\big[\Delta A_{j \setminus i}\big] - \mathbb{E}_{i \neq j}\big[A_{j \setminus i}\big],
\ee
which measures the extent to which ablating latents for class $i$ selectively degrades the corresponding probe while leaving other probes unaffected. A high TPP score thus indicates effective disentanglement of SAE latents.

\subsubsection{RAVEL}

RAVEL \cite{chaudhary2024ravel} evaluates the ability of SAEs to disentangle features by testing whether individual latents correspond to distinct factual attributes. The dataset spans five entity types (cities, Nobel laureates, verbs, physical objects, and occupations), each with 400–800 instances and 4–6 attributes (e.g., cities have country, continent, and language), probed with 30–90 natural language and JSON prompt templates.

Evaluation proceeds in three stages: (i) filtering entity and attribute pairs that the model predicts reliably, (ii) identifying attribute and specific features using probes trained on latent representations, and (iii) computing a disentanglement score that averages \emph{cause} and \emph{isolation} metrics. The \emph{cause} score measures whether intervening on a feature for attribute $A$ (e.g., setting Paris’s country to Japan) correctly changes the prediction of $A$, while the \emph{isolation} score verifies that other attributes $B$ (e.g., language = French) remain unaffected. A higher final score indicates stronger disentanglement of features.

\subsubsection{Probing Evaluation}
We assess whether SAEs capture interpretable features through targeted probing tasks across five domains: profession classification, sentiment and product categorization , language identification, programming language classification, and topic categorization. Each dataset is partitioned into multiple binary classification tasks, yielding a total of 35 evaluation tasks.

For each task, we encode inputs with the SAE, apply mean pooling over non-padding tokens, and select the topk latents via maximum mean difference. A logistic regression probe is then trained on these representations and evaluated on held-out test data. To ensure comparability across tasks, we sample 4,000 training and 1,000 test examples per task, truncate inputs to 128 tokens, and, for GitHub, exclude the first 150 characters following \cite{gurnee2023finding}. We also compare mean and max pooling, finding mean pooling slightly superior. Datasets with more than two classes are subsampled into balanced binary subsets while maintaining a positive class ratio of at least 0.2.

\subsection{Task Performance}

As shown in Table~\ref{tab:task-performance}, we find that the AbsTopK methodology exhibits a superior level of performance relative to the comparative TopK and JumpReLU techniques across the evaluated models and layers.

In particular, the AbsTopK operator performs best on the majority of the evaluation metrics. While its performance is more competitive in a few areas, its dominant strength in the other key areas makes it a robust and highly effective sparsity operator according to these results. The method's strength appears to be model-agnostic, showcasing its general applicability.

\section{Steering Methods for DiM and SAEs}
\label{app:steer_methods}
In this section, we present methods for controlling specific concepts in model representations.
For DiM, we introduce two intervention strategies: \emph{activation addition}, to amplify a concept's effect, and \emph{directional ablation}, to remove it from intermediate activations. For the HarmBench experiments, we specifically employ the activation addition method. Following this, we describe how similar steering can be achieved in SAEs through latent feature manipulation and ablation.
\paragraph{Activation addition.}
Given a concept vector $\vd^{(l)}$ extracted from layer $l$, we can modulate the corresponding feature via a simple linear intervention. Concretely, for a specific input, we add the vector to the layer activations with the strength $\alpha$ to shift them toward the concept activation, thereby inducing the given concept:

\begin{equation}
    \vx^{(l)\prime} \gets \alpha \vd^{(l)} + \vx^{(l)}.
\end{equation}

This intervention is applied only at layer $l$ and across all token positions.

\paragraph{Directional ablation.}
To study the role of a particular direction $\vd$ in the model’s computation, we can remove it from the representations using directional ablation. Specifically, we zero out the component along $\vd$ for every residual stream activation $\vx$:

\begin{equation}
    \vx^{(l)\prime} \gets \vx^{(l)} - \alpha\vd \vd^\top \vx^{(l)}.
\end{equation}

This operation is applied to every activation $\vx^{(l)}$, across all layers $l$, effectively preventing the model from encoding this direction in its residual stream.

\paragraph{SAE Latent feature clamping.}
For a target latent feature $z_i$ in the SAE feature vector $\vz$, we can modulate its influence on model behavior by clamping it to a constant $c \in \mathbb{R}$. Denote a feature vector $\vz$, and let $\vz_{i,c}$ be the modified vector with $z_i$ replaced by $c$.

Define the clamping function $C_{i,c}$ as

\begin{equation}
    [C_{i,c}(\vz)]_k =
    \begin{cases}
        z_k & \text{if } k \neq i, \\
        c   & \text{if } k = i,
    \end{cases}
\end{equation}

so that $C_{i,c}(\vz) = \vz_{i,c}$.

In conventional SAEs, this clamping strategy can be interpreted as a directional control: setting $c$ to a negative value suppresses the corresponding concept, while a positive $c$ encourages it. We adopt a similar approach to perform steering in our framework, using clamping to directly modulate individual latent features and thereby control the presence or absence of specific semantic concepts in the reconstructed representation.

\end{document}